\documentclass[11pt]{IEEEtran}

 \usepackage{ownstyle3}
 \usepackage{hyperref}

 \title{Bayesian regression and Bitcoin}

\author{
Devavrat Shah   \qquad Kang Zhang
\\
Laboratory for Information and Decision Systems
\\
Department of EECS
\\ 
Massachusetts Institute of Technology\\
\texttt{devavrat@mit.edu, zhangkangj@gmail.com}
}

\begin{document}

\maketitle
\begin{abstract} In this paper, we discuss the method of Bayesian regression and its efficacy for predicting price variation of Bitcoin, 
a recently popularized virtual, cryptographic currency. Bayesian regression refers to utilizing empirical data as proxy to perform 
Bayesian inference. We utilize Bayesian regression for the so-called ``latent source model''. The Bayesian regression for 
``latent source model'' was introduced and discussed by Chen, Nikolov and Shah \cite{latent1} and Bresler, Chen and 
Shah \cite{latent2} for the purpose of binary classification. They established theoretical as well as empirical efficacy of the 
method for the setting of binary classification. 

In this paper,  instead we utilize it for predicting real-valued quantity, the price of Bitcoin. Based on this price prediction method, we
devise a simple strategy for trading Bitcoin. The strategy is able to nearly double the investment in less than 60 day period when
run against real data trace. 

\end{abstract}

\section{Bayesian Regression}

\noindent
{\bf The problem.} We consider the question of regression: we are given $n$ training labeled data points $(x_i, y_i)$ for $1\leq i\leq n$ with $x_i \in \R^d, y_i \in \R$ 
for some fixed $d\geq 1$. The goal is to use this training data to predict the unknown label $y \in \R$ for given $x \in \R^d$. 

\medskip
\noindent 
{\bf The classical approach.} A standard approach from non-parametric statistics (cf. see \cite{wasserman} for example) is to assume model of the following type: the labeled data is generated in
accordance with relation $y = f(x) + \beps$ where $\beps$ is an independent random variable representing noise, usually assumed to be Gaussian 
with mean $0$ and (normalized) variance $1$. The regression method boils down to estimating $f$ from $n$ observation $(x_1,y_1), \dots, (x_n, y_n)$
and using it for future prediction.  For example, if $f(x) = x^T \theta^*$, i.e. $f$ is assumed to be linear function, then the classical least-squares estimate is
used for estimating $\theta^*$ or $f$: 
\begin{align}
\hat{\theta}_{\small LS} & \in \argmin_{\theta \in \R^d} \sum_{i=1}^n (y_i - x_i^T \theta)^2
\end{align}
In the classical setting, $d$ is assumed fixed and $n \gg d$ which leads to justification of such an estimator being highly effective. In various modern
applications, $n \asymp d$ or even $n \ll d$ is more realistic and thus leaving highly under-determined problem for estimating $\theta^*$. Under
reasonable assumption such as `sparsity' of $\theta^*$, i.e. $\|\theta^*\|_0 \ll d$, where $\|\theta^*\|_0 = |\{ i: \theta^*_i \neq 0\}|$, the regularized least-square
estimation (also known as Lasso \cite{lasso}) turns out to be the right solution: for appropriate choice of $\lambda > 0$, 
\begin{align}
\hat{\theta}_{\small LASSO} & \in \argmin_{\theta \in \R^d} \sum_{i=1}^n (y_i - x_i^T \theta)^2 + \lambda \|\theta\|_1.
\end{align}

At this stage, it is worth pointing out that the above framework, with different functional forms, has been extremely successful in practice. And very exciting 
mathematical development has accompanied this theoretical progress. The book \cite{wasserman} provides a good overview of this literature. Currently, 
it is a very active area of research. 

\medskip
\noindent 
{\bf Our approach.} The key to success for the above stated approach lies in the ability to choose a reasonable parametric function space over which one tries to
estimate parameters using observations. In various modern applications (including the one considered in this paper), making such a choice seems challenging.
The primary reason behind this is the fact that the data is very high dimensional (e.g. time-series) making either parametric space too complicated or meaningless. 
Now in many such scenarios, it seems that there are few prominent ways in which underlying event exhibits itself. For example, a phrase or collection of words
become viral on Twitter social media for few different reasons -- a public event, life changing event for a celebrity, natural catastrophe, etc. Similarly, there are
only few different types of people in terms of their choices of movies -- those of who like comedies and indie movies, those who like court-room dramas, etc. 
Such were the insights formalized in works \cite{latent1} and \cite{latent2} as the `latent source model' which we describe formally in the context of above described 
framework. 

There are $K$ distinct latent sources $s_1,\dots, s_K \in \R^d$; a latent distribution over $\{1,\dots, K\}$ with associated probabilities $\{\mu_1,\dots, \mu_K\}$; 
and $K$ latent distributions over $\R$, denoted as $\P_1, \dots, \P_K$. Each labeled data point $(x,y)$ is generated as follows. Sample index $T \in \{1,\dots, K\}$
with $\P(T = k) = \mu_k$ for $1\leq k\leq K$; $x = s_T + \beps$, where $\beps$ is $d$-dimensional independent random variable, representing noise,  
which we shall assume to be Gaussian with mean vector $\bzero = (0,...,0) \in \R^d$ and identity covariance matrix; $y$ is sampled from $\R$ 
as per distribution $\P_T$. 

Given this model, to predict label $y$ given associated observation $x$, we can utilize the conditional distribution\footnote{Here we are assuming that 
the random variables have well-defined densities over appropriate space. And when appropriate, conditional probabilities are effectively representing conditional
probability density.} of $y$ given $x$ given as follows:
\begin{align}
\P\big(y \big | x\big) & = \sum_{k=1}^T \P\big( y \big | x, T = k\big) \P\big( T=k \big | x\big) \nonumber \\
& \propto  \sum_{k=1}^T \P\big( y \big | x, T = k\big) \P\big( x \big | T=k  \big) \P(T =k)  \nonumber \\
& =  \sum_{k=1}^T \P_k \big( y \big) \P\big( \beps = (x - s_k) \big) \mu_k  \nonumber \\
& =  \sum_{k=1}^T \P_k \big( y \big) \exp\Big(-\frac{1}{2} \|x - s_k\|_2^2 \Big) \mu_k. \label{eq:1}
\end{align}

Thus, under the latent source model, the problem of regression becomes a very simple Bayesian inference problem. However, the problem is lack
of knowledge of the `latent' parameters of the source model. Specifically, lack of knowledge of $K$, sources $(s_1, \dots, s_K)$, probabilities 
$(\mu_1,\dots, \mu_K)$ and probability distributions $\P_1,\dots, \P_K$. 

To overcome this challenge, we propose the following simple algorithm: utilize empirical data as proxy for estimating conditional distribution of $y$ given
$x$ given in \eqref{eq:1}. Specifically, given $n$ data points $(x_i, y_i)$, $1\leq i\leq n$, the empirical conditional probability is 
\begin{align}
\P_{emp} \big(y \big | x\big) &  =  \frac{\sum_{i=1}^n \ind(y = y_i) \exp\Big(-\frac{1}{4} \|x - x_i\|_2^2 \Big)}{\sum_{i=1}^n \exp\Big(-\frac{1}{4} \|x - x_i\|_2^2 \Big)}. \label{eq:2}
\end{align}

The suggested empirical estimation in \eqref{eq:2} has the following implications: in the context of binary classification, $y$ takes values in $\{0,1\}$. Then 
\eqref{eq:2} suggests the following classification rule: compute ratio 
\begin{align}
& \frac{\P_{emp} \big(y=1 \big | x\big)}{\P_{emp} \big(y=0 \big | x\big)} \nonumber \\
& \quad =  \frac{\sum_{i=1}^n \ind(y_i = 1) \exp\Big(-\frac{1}{4} \|x - x_i\|_2^2 \Big)}{\sum_{i=1}^n \ind(y_i = 0) \exp\Big(-\frac{1}{4} \|x - x_i\|_2^2 \Big)}. \label{eq:3}
\end{align}
If the ratio is $> 1$, declare $y = 1$, else declare $y = 0$. In general, to estimate the conditional expectation of $y$, given observation $x$, \eqref{eq:2} suggests 
\begin{align}
\E_{emp}[y | x ] & = \frac{\sum_{i=1}^n y_i  \exp\Big(-\frac{1}{4} \|x - x_i\|_2^2 \Big)}{\sum_{i=1}^n \exp\Big(-\frac{1}{4} \|x - x_i\|_2^2 \Big)}. \label{eq:4}
\end{align}
Estimation in  \eqref{eq:4} can be viewed equivalently as a `linear' estimator: let vector $X(x) \in \R^n$ be such that $X(x)_i =  \exp\Big(-\frac{1}{4} \|x - x_i\|_2^2 \Big)/Z(x)$
with $Z(x) = \sum_{i=1}^n \exp\Big(-\frac{1}{4} \|x - x_i\|_2^2 \Big)$, and $\by  \in \R^n$ with $i$th component being $y_i$, then $\hat{y} \equiv \E_{emp}[y | x ] $ is 
\begin{align}
\hat{y}  & = X(x) \by. \label{eq:5} 
\end{align}
In this paper, we shall utilize \eqref{eq:5} for predicting future variation in the price of Bitcoin. This will further feed into a trading strategy. The details are discussed
in the Section \ref{sec:bitcoin}.

\medskip
\noindent
{\bf Related prior work.} To begin with, Bayesian inference is foundational and use of empirical data as a proxy has been a well known
approach that is potentially discovered and re-discovered in variety of contexts over decades, if not for centuries. For example, \cite{bishop2003bayesian} provides a nice
overview of such a method for a specific setting (including classification).  The concrete form \eqref{eq:2} that results due to the assumption of
latent source model is closely related to the popular rule called the `weighted majority voting' in the literature. It's asymptotic effectiveness is
discussed in literature as well, for example \cite{fukunaga1990introduction}. 

The utilization of latent source model for the purpose of identifying precise sample complexity for Bayesian regression was first studied in \cite{latent1}.
In \cite{latent1}, authors showed the efficacy of such an approach for predicting trends in social media Twitter. For the purpose of the specific application,
authors had to utilize noise model that was different than Gaussian leading to minor change in the \eqref{eq:2} -- instead of using quadratic function, it was
quadratic function applied to logarithm (component-wise) of the underlying vectors - see \cite{latent1} for further details.  

In various modern application such as online recommendations, the observations ($x_i$ in above formalism) are only partially observed. This
requires further modification of \eqref{eq:2} to make it effective. Such a modification was suggested in \cite{latent2} and corresponding theoretical
guarantees for sample complexity were provided. 

We note that in both of the works \cite{latent1, latent2}, the Bayesian regression for latent source model was used primarily for binary classification. 
Instead, in this work we shall utilize it for estimating real-valued variable. 

\section{Trading Bitcoin}
\label{sec:bitcoin}

\noindent
{\bf What is Bitcoin.} Bitcoin is a peer-to-peer cryptographic digital currency that was created in 2009 by an unknown person using the alias Satoshi Nakamoto 
\cite{bitcoin-info1,bitcoin-info2}. 
Bitcoin is unregulated and hence comes with benefits (and potentially a lot of issues) such as transactions can be done in a frictionless manner - no fees - and
anonymously. It can be purchased through exchanges or can be `mined' by computing/solving complex mathematical/cryptographic puzzles. Currently, 25
Bitcoins are rewarded every 10 minutes (each valued at around US \$400 on September 27, 2014).  As of September 2014, its daily transaction volume is in 
the range of US \$30-\$50 million and its market capitalization has exceeded US \$7 billion. With such huge trading volume, it makes sense to think of it as a 
proper financial instrument as part of any reasonable  quantitative (or for that matter any) trading strategy. 

In this paper, our interest is in understanding whether there is `information' in the historical data related to Bitcoin that can help predict future price variation in the 
Bitcoin and thus help develop profitable quantitative strategy using Bitcoin. As mentioned earlier, we shall utilize Bayesian regression inspired by latent source 
model for this purpose. 

\medskip
\noindent
{\bf Relevance of Latent Source Model.} Quantitative trading strategies have been extensively studied and applied in the financial industry, although many of 
them are kept secretive. One common approach reported in the literature is technical analysis, which assumes that price movements follow a set of patterns and 
one can use past price movements to predict future returns to some extent \cite{lo88, lo99}. Caginalp and Balenovich  \cite{cag03} showed that some patterns emerge from 
a model involving two distinct groups of traders with different assessments of valuation. Studies found that some empirically developed geometric 
patterns, such as heads-and-shoulders, triangle and double-top-and-bottom, can be used to predict future price changes \cite{lo00, cag88, park04}. 

The Latent Source Model is precisely trying to model existence of such underlying patterns leading to price variation. Trying to develop patterns with the help
of a human expert or trying to identify patterns explicitly in the data,  can be challenging and to some extent subjective. Instead, using Bayesian
regression approach as outlined above allows us to utilize the existence of patterns for the purpose of better prediction without explicitly finding them. 

\medskip
\noindent
{\bf Data.} 
%Bitcoin data was collected about eight exchanges using publicly available APIs between February 2014-July 2014. The
%data included information about price, order book and trade history. The total data points were over 200 million. 
In this paper, to perform experiments, we have used data related to price and order book obtained from {\tt Okcoin.com} -- one of the largest 
exchanges operating in China. The data concerns time period between February 2014 to July 2014. The total raw data points were
over 200 million.  The order book data consists of 60 best prices at which one is
willing to buy or sell at a given point of time. The data points were acquired at the interval of every two seconds. For the purpose
of computational ease, we constructed a new time series with time interval of length 10 seconds; each of the raw data point was
mapped to the closest (future) 10 second point. While this coarsening introduces slight `error' in the accuracy, since our trading 
strategy operates at a larger time scale, this is insignificant.

\medskip
\noindent
{\bf Trading Strategy.} The trading strategy is very simple: at each time, we either maintain position of $+1$ Bitcoin, $0$ Bitcoin or $-1$ Bitcoin. At 
each time instance, we predict the average price movement over the 10 seconds interval, say $\Delta p$, using Bayesian regression (precise details explained below) - 
if $\Delta p > t$, a threshold, then we buy a bitcoin if current bitcoin position is $\leq 0$; if $\Delta p < -t$, then we sell a bitcoin if current position is $\geq 0$; else
do nothing. The choice of time steps when we make trading decisions as mentioned above are chosen carefully by looking at the recent trends. We skip 
details as they do not have first order effect on the performance. 

\medskip
\noindent
{\bf Predicting Price Change.} The core method for average price change $\Delta p$ over the 10 second interval is the Bayesian regression as in \eqref{eq:5}. 
Given time-series of price variation of Bitcoin over the interval of few months, measured every 10 second interval, we have a very large time-series (or a 
vector). We use this historic time series and from it, generate three subsets of time-series data of three different lengths: $S_1$ of time-length 30 minutes, 
$S_2$ of time-length 60 minutes, and $S_3$ of time-length 120 minutes. Now at a given point of time, to predict the future change $\Delta p$, we use the
historical data of three length: previous 30 minutes, 60 minutes and 120 minutes - denoted $x^1, x^2$ and $x^3$. We use $x^j$ with historical samples $S^j$ 
for Bayesian regression (as in \eqref{eq:5}) to predict average price change $\Delta p^j$ for $1\leq j \leq 3$. We also calculate 
$r = (v_{bid} - v_{ask})/(v_{bid} + v_{ask})$  where $v_{bid}$ is total volume people are willing to buy in the top 60 orders and 
$v_{ask}$ is the total volume people are willing to sell in the top 60 orders based on the current order book data. The final estimation 
$\Delta p$ is produced as 
\begin{align}
\Delta p & = w_0 + \sum_{j=1}^3 w_j \Delta p^j + w_4 r, 
\end{align}
where $\bw = (w_0, \dots, w_4)$ are learnt parameters.  In what follows, we explain how $S_j,~1\leq j\leq 3$ are collected; and how $\bw$ is learnt. This will complete the description of the price change prediction 
algorithm as well as trading strategy. 

Now on finding $S_j, 1\leq j\leq 3$ and learning $\bw$. We divide the entire time duration into three, roughly equal sized, periods. We utilize the first time period to find patterns $S_j, ~1\leq j \leq 3$. The second period 
is used to learn parameters $\bw$ and the last third period is used to evaluate the performance of the algorithm. The learning of $\bw$ is done simply by finding the best linear fit over all choices given
the selection of $S_j, ~1\leq j\leq 3$. Now selection of $S_j, ~1\leq j\leq 3$. For this, we take all possible time series of appropriate length (effectively vectors of dimension $180$, $360$ and $720$ respectively
for $S_1, S_2$ and $S_3$). Each of these form $x_i$ (in the notation of formalism used to describe \eqref{eq:5}) and their corresponding label $y_i$ is computed by looking at the average price change in
the $10$ second time interval following the end of time duration of $x_i$. This data repository is extremely large. {To facilitate computation on single machine with $128G$ RAM with $32$ cores, we clustered
patterns in 100 clusters using $k-$means algorithm. From these, we chose 20 most effective clusters and took representative patterns from these clusters. 

The one missing detail is computing `distance' between pattern $x$ and $x_i$ - as stated in \eqref{eq:5}, this is squared $\ell_2$-norm. Computing $\ell_2$-norm is computationally intensive. 
For faster computation, we use negative of `similarity', defined below, between patterns as 'distance'. 
\begin{definition}(Similarity)
The similarity between two vectors $\a$, $\mathbf{b} \in \R^M$ is defined as 
\begin{align}
s(\a, \mathbf{b}) & = \frac{\sum_{z =1}^M (a_z - mean(\a)) (b_z - mean(\mathbf{b}))}{M ~std(\a)~ std(\mathbf{b})},
\end{align}
where $mean(\a) = (\sum_{z=1}^M a_z)/M$ (respectively for $\mathbf{b}$) and $std(\a) = (\sum_{z=1}^M (a_i - mean(\a))^2)/M$ (respectively for $\mathbf{b}$).
\end{definition}
In \eqref{eq:5}, we use $\exp(c \cdot s(x, x_i))$ in place of $\exp(-\|x - x_i\|_2^2/4)$ with choice of constant $c$ optimized for better prediction using the fitting data (like for $\bw$). 

We make a note of the fact that this similarity can be computed very efficiently by storing the pre-computed patterns (in $S_1, S_2$ and $S_3$) in a normalized form 
($0$ mean and std $1$). In that case, effectively the computation boils down to performing an inner-product of vectors, which can be done very efficiently. For example, 
using a straightforward Python implementation, more than 10 million cross-correlations can be computed in 1 second using 32 core machine with $128G$ RAM.

\begin{figure}[ht]
\vskip 0.2in
\begin{center}
\centerline{\includegraphics[width=\columnwidth]{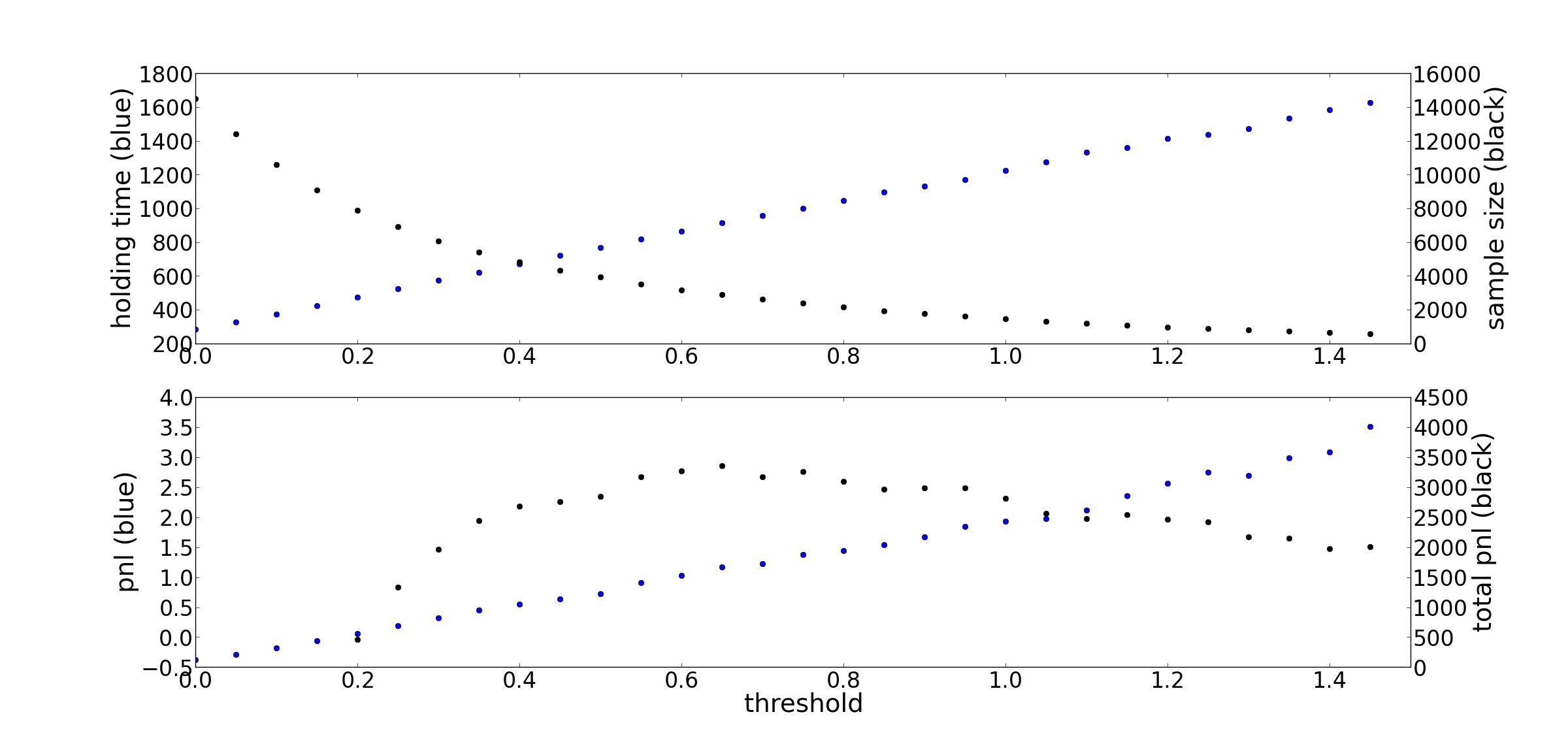}}
\caption{The effect of different threshold on the number of trades, average holding time and profit}
\label{strategy1}
\end{center}
\vskip -0.2in
\end{figure} 

\begin{figure}[ht]
\vskip 0.2in
\begin{center}
\centerline{\includegraphics[width=\columnwidth]{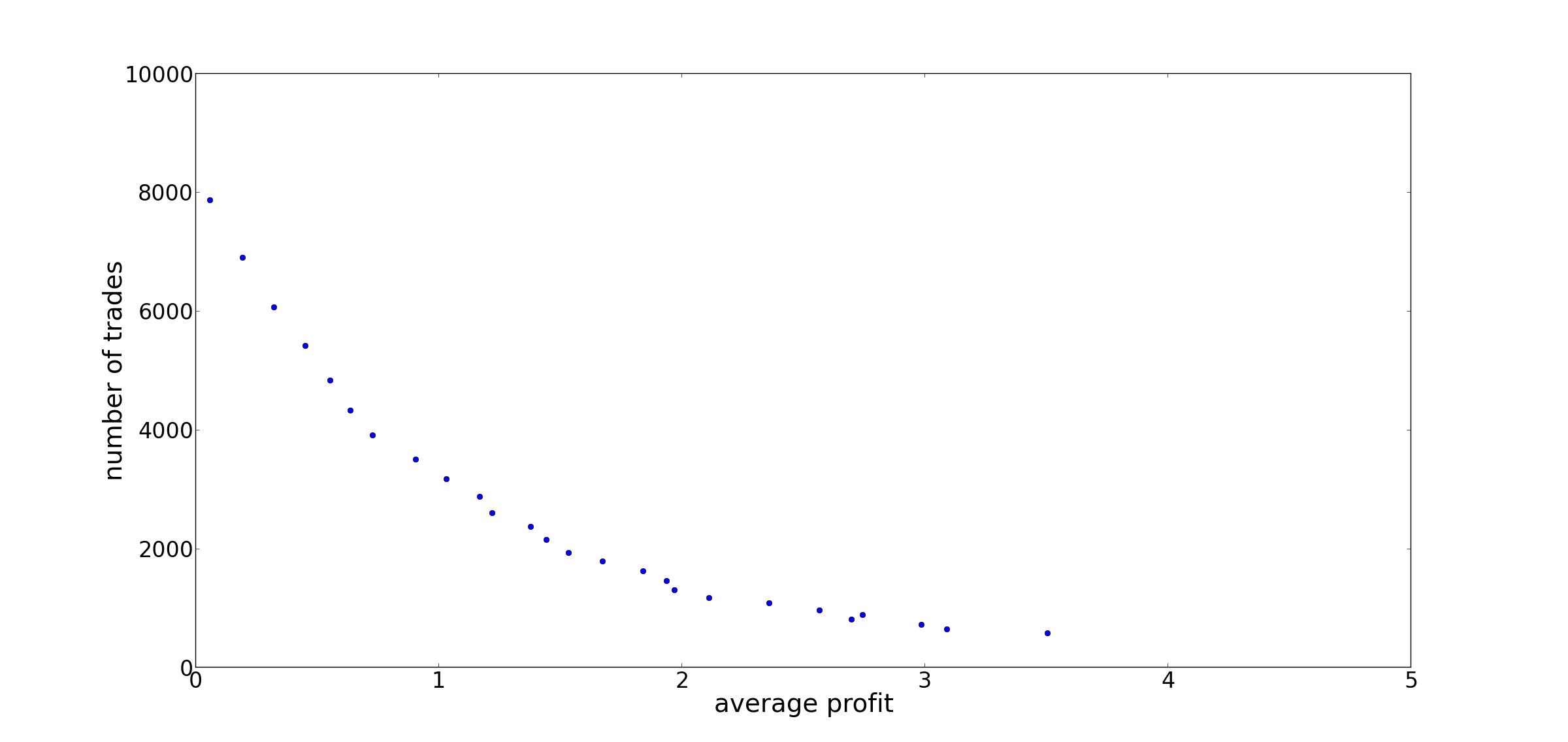}}
\caption{The inverse relationship between the average profit per trade and the number of trades}
\label{strategy2}
\end{center}
\vskip -0.2in
\end{figure}

\begin{figure}[ht]
\vskip 0.2in
\begin{center}
\centerline{\includegraphics[width=\columnwidth]{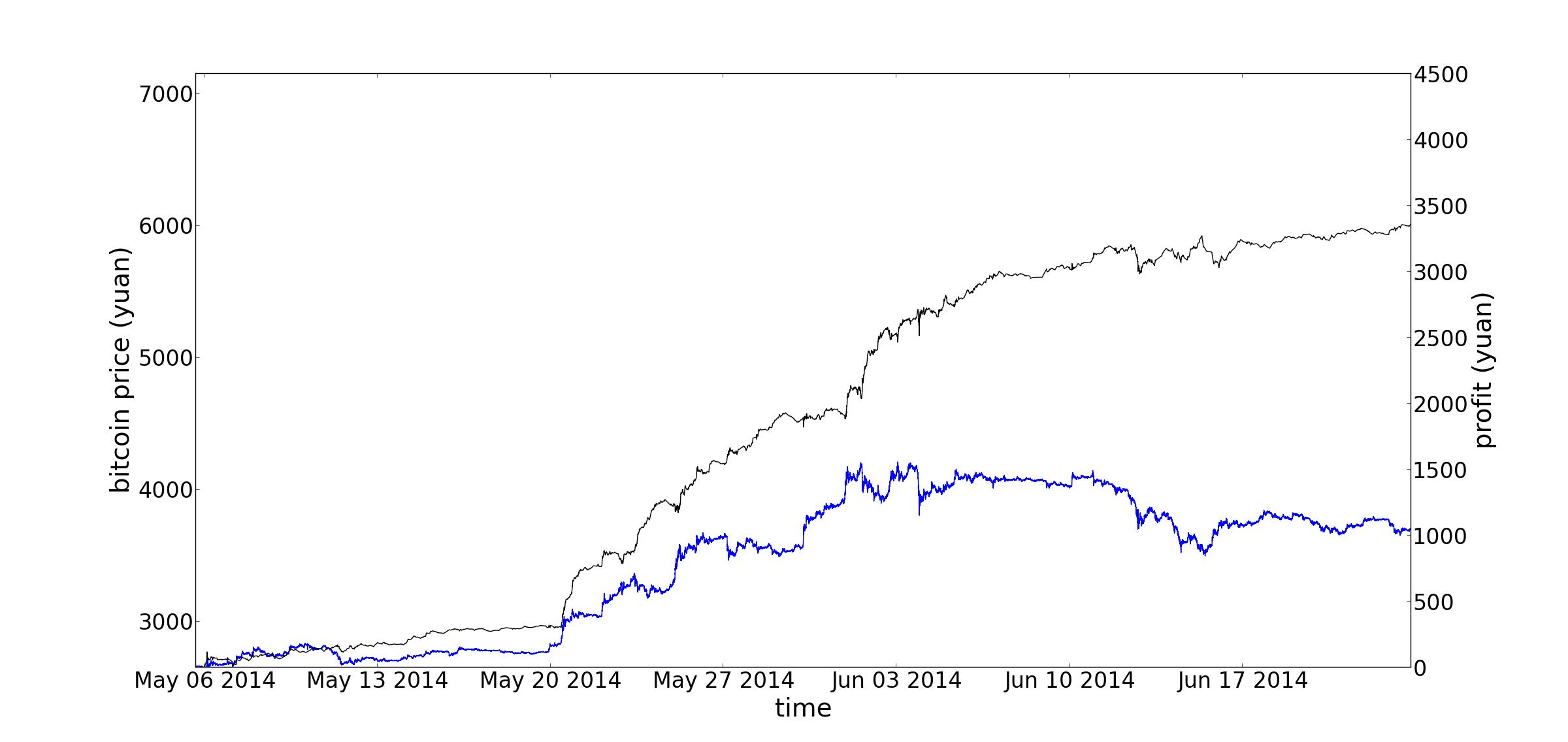}}
\caption{The figure plots two time-series - the cumulative profit of the
strategy starting May 6, 2014 and the price of Bitcoin. The one, that is
lower (in blue), corresponds to the price of Bitcoin, while the other
corresponds to cumulative profit. The scale of $Y$-axis on left corresponds
to price, while the scale of $Y$-axis on the right corresponds to cumulative
profit. }
\label{simulation}
\end{center}
\vskip -0.2in
\end{figure}

\medskip
\noindent
{\bf Results.} We simulate the trading strategy described above on a third of total data in the duration of May 6, 2014 to June 24, 2014 in a causal manner to see how well our strategy does. The training data utilized is all historical (i.e. collected before May 6, 2014). We use different threshold $t$ and see how the performance of strategy changes.  As shown in Figure \ref{strategy1} and \ref{strategy2} different threshold provide different performance. Concretely, as we increase the threshold, the number of trades decreases and the average holding time increases. At the same time, the average profit per trade increases. 

We find that the total profit peaked at $3362$ yuan with a $2872$ trades in total with average investment of $3781$ yuan. This is roughly $89\%$ return in $50$ days with a Sharpe ratio of $4.10$. To recall, sharp ratio of strategy, over a given time period, is defined as follows: let $L$ be the number of trades made during the time interval; let $p_1,\dots, p_L$ be the profits (or losses if they are negative valued) made in each of these trade; let 
$C$ be the modulus of difference between start and end price for this time interval, then Sharpe ratio \cite{sharpe} is
\begin{align}
\frac{\sum_{\ell=1}^L p_\ell - C}{L \sigma_p},
\end{align}
where $\sigma_p = \frac{1}{L}\Big(\sum_{\ell=1}^L (p_\ell - \bar{p})^2)$ with $\bar{p} = (\sum_{\ell=1}^L p_\ell)/L$. Effectively, Sharpe ratio of a strategy captures how well the strategy performs compared to the 
risk-free strategy as well as how consistently it performs. 

Figure \ref{simulation} shows the performance of the best strategy over time. Notably, the strategy performs better in the middle section when the market volatility is high. In addition, the strategy is still profitable even when the price is decreasing in the last part of the testing period. 

\section{Discussion}

\medskip
\noindent
{\bf Are There Interesting Patterns?} The patterns utilized in prediction were clustered using the standard $k-$means algorithm. The clusters (with high price variation, and confidence) were carefully inspected. The cluster centers (means as defined by $k-$means algorithm) found are reported in Figure \ref{triangle}. As can be seen, there are ``triangle'' pattern and ``head-and-shoulder'' pattern. Such patterns are observed and reported in the technical analysis literature. This seem to suggest that there are indeed such patterns and provides evidence of the existence of latent source model and explanation of success of our trading strategy. 

\begin{figure}[ht]
\vskip 0.2in
\begin{center}
\centerline{\includegraphics[width=\columnwidth]{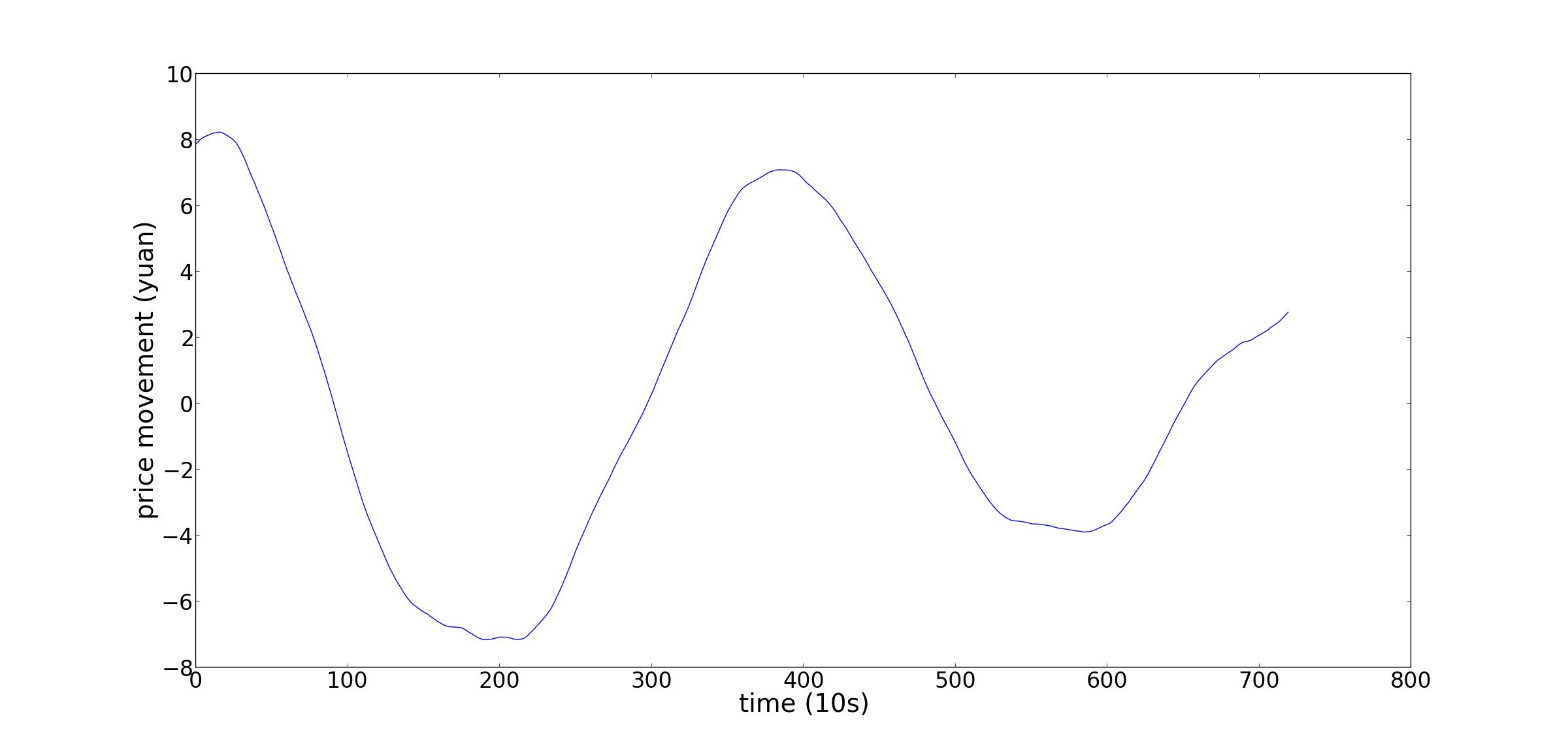}}
\centerline{\includegraphics[width=\columnwidth]{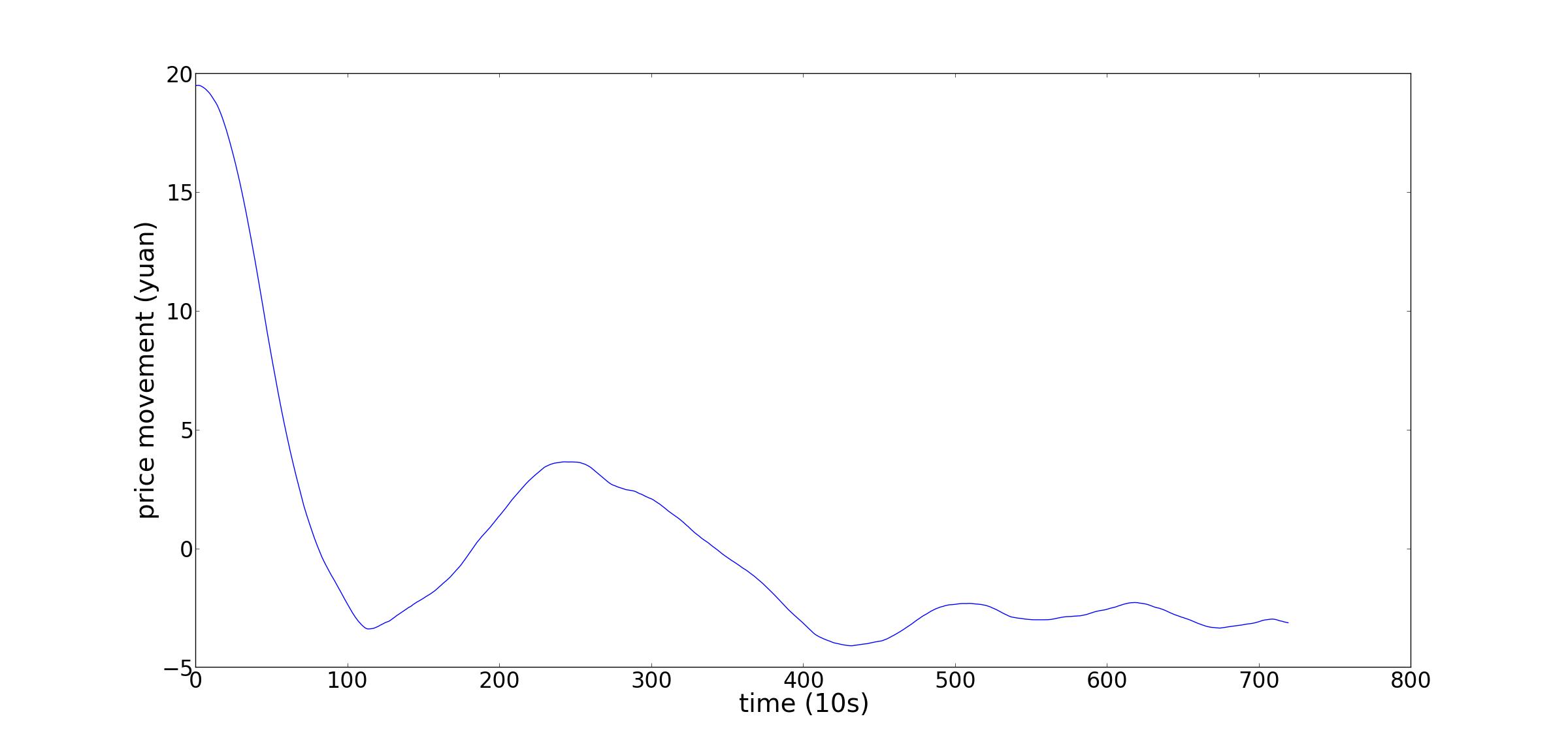}}
\caption{Patterns identified resemble the  head-n-shoulder and triangle pattern as shown above.}
\label{triangle}
\end{center}
\vskip -0.2in
\end{figure}

\medskip
\noindent
{\bf Scaling of Strategy.} The strategy experimented in this paper holds minimal position - at most $1$ Bitcoin ($+$ or $-$). This leads to nearly doubling of investment in 50 days. Natural question arises - does this scale for large volume of investment? Clearly, the number of transactions at a given price at a given instance will decrease given that order book is always finite, and hence linearity of scale is not expected. On the other hand, if we allow for flexibility in the position (i.e more than $\pm 1$), then it is likely that more profit can be earned. Therefore, to scale such a strategy further careful research is required. 

\medskip
\noindent
{\bf Scaling of Computation.} To be computationally feasible, we utilized `representative' prior time-series. It is definitely believable that using all possible time-series could have improved the prediction power and hence the efficacy of the strategy. However, this requires computation at massive scale. Building a scalable computation architecture is feasible, in principle as by design \eqref{eq:5} is trivially parallelizable (and map-reducable) computation. Understanding the role of computation in improving prediction quality remains important direction for investigation.  

\section*{Acknowledgments}
This work was supported in part by NSF grants CMMI-1335155 and CNS-1161964, and by Army Research Office MURI Award W911NF-11-1-0036.

\bibliographystyle{ieeetr}
\bibliography{bibliography}

\end{document}